%% file: icra.tex
\documentclass[letterpaper, 10 pt, conference]{ieeeconf}  

\pdfoutput=1
\IEEEoverridecommandlockouts                              

\usepackage{xcolor}
\usepackage{amsmath} 
\usepackage{amsfonts}
\usepackage{graphicx}
\usepackage{amsmath}
\usepackage{hyperref}
\usepackage{tabularx}
\usepackage{booktabs}
\usepackage{multirow}
\let\labelindent\relax
\usepackage{enumitem}

\title{\LARGE \bf
4DRadar-GS: Self-Supervised Dynamic Driving Scene Reconstruction \\with 4D Radar
}

\author{Xiao Tang\textsuperscript{\rm 1}, Guirong Zhuo\textsuperscript{\rm 1}, Cong Wang\textsuperscript{\rm 2}, Boyuan Zheng\textsuperscript{\rm 1},\\ Minqing Huang\textsuperscript{\rm 1}, Lianqing Zheng\textsuperscript{\rm 1}, Long Chen\textsuperscript{\rm 2}, Shouyi Lu\textsuperscript{\rm 1}$^{*}$ 
\thanks{* Corresponding author.}
\thanks{This work was supported by the National Natural Science Foundation of China under Grant 52325212, the National Key Research and Development Program of China (No. 2022YFE0117100), Shanghai Tongyu Automobile Technology Intelligent Vehicle By-Wire Chassis Joint Laboratory}
\thanks{$^{1}$Xiao Tang, Shouyi Lu, Guirong Zhuo, Minqing Huang, Boyuan Zheng and Lianqing Zheng are with the School of Automotive Studies, Tongji University, Shanghai, China.}%
\thanks{$^{2}$Cong Wang and Long Chen are with the State Key Laboratory of Multimodal Artificial Intelligence Systems, Institute of Automation, Chinese Academy of Sciences.}%
}

\begin{document}

\maketitle
\thispagestyle{empty}
\pagestyle{empty}

\begin{abstract}

3D reconstruction and novel view synthesis are critical for validating autonomous driving systems and training advanced perception models. Recent self-supervised methods have gained significant attention due to their cost-effectiveness and enhanced generalization in scenarios where annotated bounding boxes are unavailable. However, existing approaches, which often rely on frequency-domain decoupling or optical flow, struggle to accurately reconstruct dynamic objects due to imprecise motion estimation and weak temporal consistency, resulting in incomplete or distorted representations of dynamic scene elements. To address these challenges, we propose 4DRadar-GS, a 4D Radar-augmented self-supervised 3D reconstruction framework tailored for dynamic driving scenes. Specifically, we first present a 4D Radar-assisted Gaussian initialization scheme that leverages 4D Radar's velocity and spatial information to segment dynamic objects and recover monocular depth scale, generating accurate Gaussian point representations. In addition, we propose a Velocity-guided PointTrack (VGPT) model, which is jointly trained with the reconstruction pipeline under scene flow supervision, to track fine-grained dynamic trajectories and construct temporally consistent representations. Evaluated on the OmniHD-Scenes dataset, 4DRadar-GS achieves state-of-the-art performance in dynamic driving scene 3D reconstruction. 

\end{abstract}


\input{sec/01_intro}

\input{sec/02_related}

\input{sec/03_method}

\input{sec/04_exp}

\input{sec/05_conclu}








\bibliographystyle{IEEEtran}
\bibliography{icra}

\end{document}

%% file: sec/01_intro.tex
\section{INTRODUCTION}

Closed-loop simulation of driving scenarios is of significant importance for the testing of autonomous driving functions and the training of large driving models \cite{dosovitskiy2017carla,li2025perception,xiong2025drivinggaussian++}. Recent advancements in 3D reconstruction and novel view synthesis techniques have introduced new research opportunities in the field of closed-loop simulation. Starting from 2D images of real-world scenes, these 3D reconstruction methods enable the reconstruction of realistic street views. This approach can substantially reduce the sim-to-real gap compared to simulators based on virtual engines.

3D Gaussian Splatting \cite{kerbl20233d} is an efficient method for 3D reconstruction and novel view synthesis, utilizing Gaussian ellipsoids to represent the geometric structures of a 3D scene. It has demonstrated robust performance in reconstructing static indoor environments or objects. However, driving scenarios pose greater challenges due to their high dynamism, incorporating moving vehicles and pedestrians. To enhance the dynamic reconstruction of driving scenes, some approaches have proposed leveraging LiDAR to provide initial point cloud inputs and using 3D bounding boxes \cite{yan2024street, chen2024omnire, wang2025drivesplat} to decouple moving actors from the scene for separate modeling. The dynamic-static decoupling strategy effectively mitigates motion blur issues. Nonetheless, obtaining accurate bounding boxes is challenging and costly, prompting exploration into self-supervised dynamic-static decoupling reconstruction methods. These methods attempt to reconstruct scenes using frequency domain decoupling \cite{chen2023periodic, peng2025desire}, semantic masks, or estimated optical flow \cite{sun2025splatflow} to enable dynamic driving scene reconstruction without requiring bounding boxes. However, these methods suffer from false or incomplete detections of dynamic targets (e.g., misclassifying stationary roadside vehicles as dynamic), which hinders effective dynamic-static decoupling. Furthermore, their motion modeling approach for dynamic points struggles to establish accurate inter-frame correspondence for dynamic actors, particularly in the presence of rapid ego or actor motion, leading to a suboptimal reconstruction of the dynamic elements within the scene.

Recently, 4D Radar has garnered significant attention from both academia and industry due to its exceptional capabilities in dynamic object perception. Leveraging its precise spatial localization and velocity measurement abilities, 4D Radar can effectively capture the motion information of dynamic objects. This is particularly advantageous in complex driving scenarios, where it can overcome the limitations faced by traditional sensors, such as LiDAR and cameras, in highly dynamic environments.

To address these challenges, we propose 4DRadar-GS, a dynamic reconstruction method assisted by 4D Radar. 
For the first challenge, we propose a 3D perception initialization scheme for the deformation field, used to directly learn a deformation field dedicated to dynamic objects from monocular driving videos. Specifically, we first utilize the dynamic perception capabilities of 4D Radar to propose a dynamic segmentation model, to accurately decouple the scene into a static background and a dynamic foreground. Secondly, utilizing the spatial localization capabilities of 4D Radar, we recover the scale of monocular depth estimation, and use the dynamic depth points as the initialization for the dynamic foreground. 
For the second challenge, we train a Velocity-guided PointTrack (VGPT) model to associate dynamic points across multiple frames. Existing methods often depend solely on two-dimensional flow estimation for supervision, which may deteriorate results due to inadequate multi-view consistency. By leveraging the radial velocity information from 4D Radar, we provide the VGPT network with additional radial supervision alongside optical flow supervision, significantly aiding the learning of subsequent four-dimensional representations and enhancing the rendering quality of dynamic actors. 
Additionally, to address frame-to-frame inconsistencies in point cloud initialization, we devised a simple yet powerful regularization technique. During training, we randomly remove Gaussian points, allowing Gaussians that are obscured by nearby Gaussian points to receive attention under sparse view conditions, thereby alleviating the issue of overfitting to training perspectives.

In summary, our key contributions can be listed as follows:
\begin{itemize}[left=0pt]
    \item We propose a novel self-supervised reconstruction framework, the first to systematically leverage 4D Radar to perform accurate dynamic decoupling and scale recovery, thereby achieving robust initialization for dynamic driving scenes.
    \item We introduce a VGPT model to establish robust temporal correspondence for dynamic actors, training it with a dual-supervision scheme that complements optical flow with direct physical constraint from 4D Radar's radial velocity.
    \item We devised a regularization method to mitigate the issue of overfitting to training viewpoints. Our approach achieves SOTA performance on the OmniHD-Scenes dataset, which was selected due to its essential 4D Radar data.
\end{itemize}

%% file: sec/02_related.tex
\section{RELATED WORKS}
\subsection{Static Scene Reconstruction with 3DGS}
3D Gaussian Splatting \cite{kerbl20233d} has emerged as a leading method for high-fidelity novel view synthesis, using millions of Gaussian primitives to achieve real-time rendering of static scenes. Following its success, initial works focused on improving its efficiency and scalability for large-scale static environments. For instance, GaussianPro \cite{cheng2024gaussianpro} introduced a progressive training scheme to enhance quality, while Octree-GS \cite{ren2024octree} employed an octree-based structure to manage Gaussians more efficiently. Subsequent works have extended 3DGS to large-scale urban environments; for instance, Hierarchical-GS \cite{kerbl2024hierarchical} achieves this by incorporating a hierarchical structure. Other approaches further optimize for large scenes by dividing the point cloud into cells and introducing Level-of-Detail representations \cite{lin2024vastgaussian, liu2025citygaussian}.
\subsection{Dynamic Scene Reconstruction with 3DGS}
A significant research thrust involves extending 3DGS from static to dynamic scenes, typically by employing deformation fields to model temporal changes, as seen in methods like Deformable-GS \cite{yang2024deformable} and 4D-GS \cite{wu20234d}. In the challenging context of autonomous driving, these techniques have been successfully applied, often with the aid of manual supervision. StreetGS \cite{yan2024street} and DrivingGaussian \cite{zhou2024drivinggaussian} first proposed a composite dynamic Gaussian framework to model multiple rigid moving objects, while OmniRe \cite{chen2024omnire} extended this capability to non-rigid pedestrians using SMPL models \cite{loper2015smpl}. A critical drawback of these methods is their heavy reliance on extensive manual annotations, which causes their reconstruction performance to degrade significantly in the presence of imprecise labels, thereby challenging their widespread application in-the-wild scenarios. 
\subsection{Self-Supervised Scene Rendering}
To eliminate this dependency on manual labels, a more challenging yet highly generalizable paradigm of self-supervised reconstruction has been explored. S3Gaussian \cite{huang2024textit} implicitly models object trajectories through a spatio-temporal decomposition network and jointly optimizes the static background and dynamic objects using various self-supervised signals, achieving photorealistic reconstruction of dynamic urban scenes in an annotation-free manner. Meanwhile, PVG \cite{chen2023periodic} proposed a unified model of Periodically Vibrating Gaussians to represent diverse objects and elements, constructing long-term trajectories by linking segments that exhibit periodic motion. DeSiRe-GS \cite{peng2025desire} extracts 2D motion masks by exploiting the poor reconstruction quality of standard 3D Gaussians in dynamic regions and introduces a temporal cross-view consistency constraint. Nevertheless, these methods often fail to establish continuous temporal correspondence for moving objects, particularly during rapid motion, leading to severe artifacts in novel view synthesis. This is a key problem that our work aims to address by introducing novel physical priors from 4D Radar.

%% file: sec/03_method.tex
\begin{figure*}[t]
\centering
\includegraphics[width=1.0\textwidth]{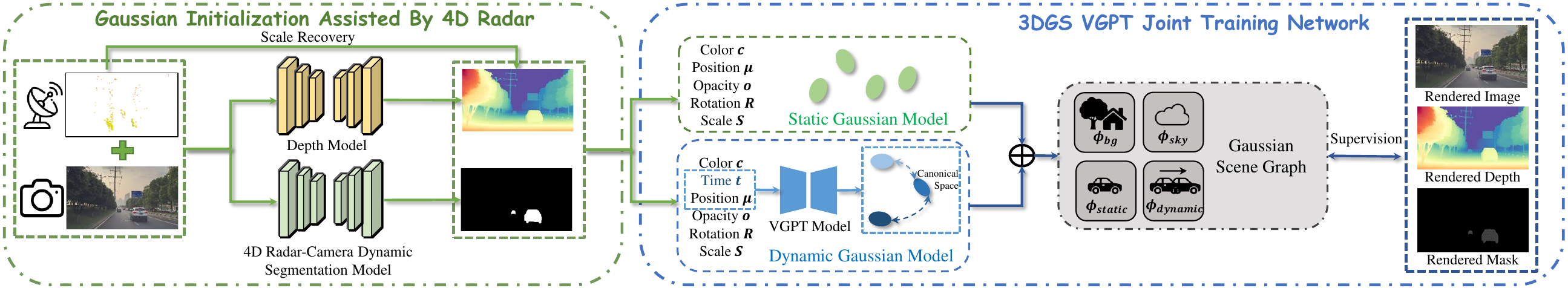}
\caption{Overview of 4DRadar-GS Framework. Our pipeline consists of two main stages. (Left) Gaussian Initialization Assisted by 4D Radar. This stage generates initial Gaussian points from 4D Radar-corrected depth and separates them into static and dynamic components using a 4D Radar-guided segmentation mask. (Right) 3DGS VGPT Joint Training Network. Here, a VGPT model maps dynamic Gaussians to a canonical space to model their motion. These, along with static Gaussians, form a complete Gaussian scene graph, jointly optimized for high-fidelity rendering, depth, and mask outputs.}
\vspace{-0.3cm}
\label{pipline_all}
\end{figure*}
\section{METHODS}
\label{sec:methods}
Our proposed framework, 4DRadar-GS, reconstructs dynamic driving scenes through the two-stage pipeline illustrated in Figure \ref{pipline_all}. We first detail our 4D Radar-assisted initialization strategy in Section \ref{subsec:radar_initialization}, where we leverage the dynamic perception and spatial localization capabilities of 4D Radar to decouple the scene into static and dynamic Gaussian primitives. We then describe the joint training network in Section \ref{subsec:joint training}, which models the motion of these dynamic primitives using a VGPT model. Finally, the complete set of objective functions and regularization techniques used for optimization are presented in Section \ref{subsec:Training and Optimization Strategies}.
\subsection{4D Radar-Assisted Gaussian Initialization}
\label{subsec:radar_initialization}
For the self-supervised reconstruction of autonomous driving scenarios, we generate depth map and dynamic mask by leveraging the dynamic perception capabilities of 4D Radar for dynamic segmentation, and its spatial localization capabilities for scale recovery. This process enables efficient partitioning of all Gaussians into two categories: object Gaussians \(\varPhi_{obj}\) and background Gaussians \(\varPhi_{bkg}\).

\textbf{4D Radar-Camera Dynamic Segmentation Model}.
\begin{figure}[t]
    \centering
    
    \includegraphics[width=0.5\textwidth]{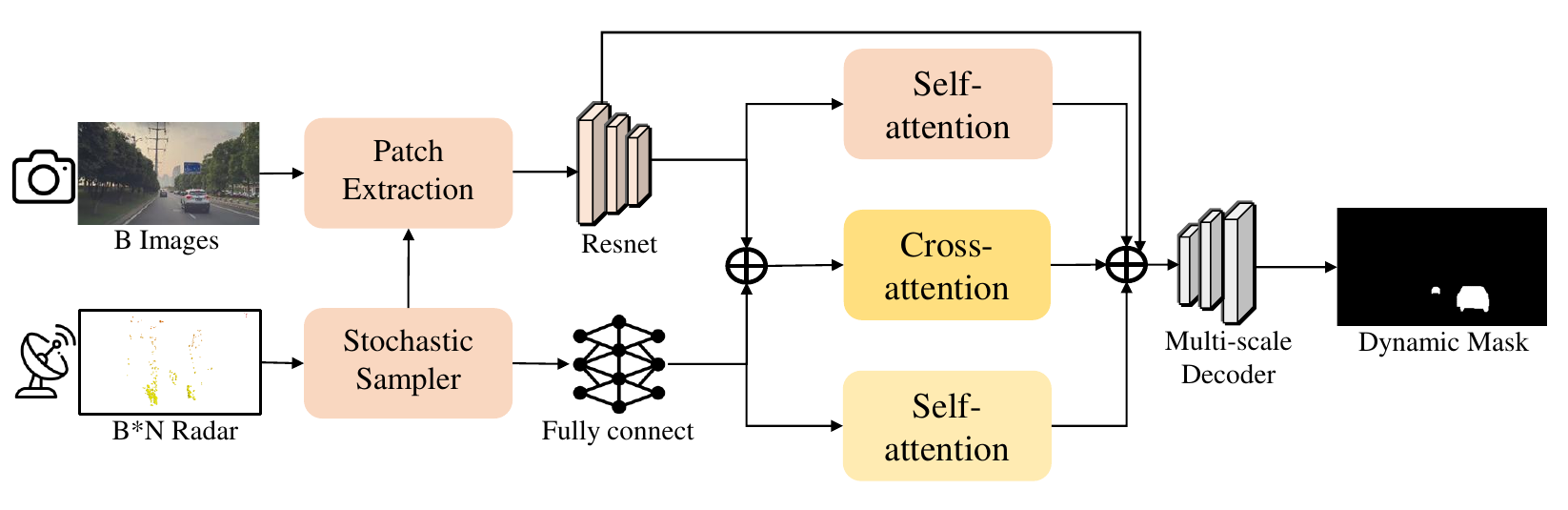}
    \vspace{-0.5cm}
    \caption{Architecture of the 4D Radar-Camera Dynamic Segmentation Model. The model fuses features from image patches and 4D Radar points using an attention-based mechanism to produce a high-resolution dynamic object mask.}
    \vspace{-0.3cm}
    \label{segment_pipline}
\end{figure}
Unlike methods based on LiDAR or SfM which struggle to directly decouple dynamic elements, 4D Radar offers the inherent advantage of providing velocity information. We leverage this by employing a 4D Radar-assisted initialization scheme that identifies dynamic objects efficiently without requiring annotated bounding box information. 

As illustrated in Figure \ref{segment_pipline}, our model predicts a dynamic segmentation mask $\hat M \in R^{H\times W}_+$ by fusing a single RGB image $I\in\mathbb R^{3\times H\times W}$ with its corresponding 4D Radar point cloud $P=\{p_i|p_i\in\mathbb R^4,i=0,1,2,\cdots,k-1\}$. First, we estimate and compensate for the ego-vehicle's motion using the RANSAC algorithm \cite{lu2025tdfanet}, enabling the precise identification of dynamic 4D Radar points. From these dynamic points, we first randomly sample a fixed number to serve as anchors. These anchors are then projected onto the image plane, and for each projected point, we define its Region of Interest (ROI) by extracting a square image patch $C_i\in\mathbb R^{3\times h\times w}$, centered at that location. These patches are subsequently processed by a ResNet \cite{he2016deep} backbone to extract multi-scale image features. Concurrently, the features of the 4D Radar points associated with each image patch, specifically their position and velocity attributes, are mean-pooled into a single vector. This vector is then processed through a fully-connected layer to match the dimensionality of the corresponding image features. The features from both modalities are then deeply fused through several layers of self-attention and cross-attention modules \cite{sun2021loftr}. Finally, the fused features are passed to a U-Net-style decoder \cite{ronneberger2015u} to generate a high-resolution dynamic probability map for each ROI, denoted as \(\hat{y}_i = h_{\theta } (C_i, p_i) \in [0, 1]^{h\times w}\).
Finally, the definitive dynamic mask \(\hat M \in R^{H\times W}_+\) is obtained by reassembling all predicted patches into their corresponding regions in the image and then classifying each region as dynamic or static based on the prediction scores and a predefined threshold \(\tau\). This process can be summarized as: 

\begin{equation}
    \hat M(u,v)=
    \begin{cases}
    1,\quad\text{if}\ \hat {y}_{max}(u,v)>\tau\\
    0,\quad\text{otherwise}
    \end{cases}
    \label{eq:case}
\end{equation}

The final mask is obtained by compositing the probability maps from all ROIs into a global confidence map. Where these maps overlap, the resulting confidence is determined by the maximum value across all predictions covering a given pixel. Specifically, the value for each pixel \((u,v)\) is determined by the ROI identified by \(\mathop{\arg\max}\limits_{i}\,\hat{y}_i(u,v)\).

\textbf{Monocular Depth Estimation and Scale Recovery}. Monocular depth estimation algorithms, such as DepthAnythingV2 \cite{yang2024depthv2} employed in this work, inherently recover the relative depth of a scene, thus lacking a true physical scale. To resolve this inherent scale ambiguity, we leverage 4D Radar to perform scale recovery. Our core idea is to perform cross-modal data association on the surface of a unit sphere by projecting two sets of points onto it: the visual 3D point cloud, obtained by back-projecting the single-frame depth map, and the static 4D Radar point cloud, isolated after ego-motion compensation. Given the more uniform spatial distribution of visual points on the unit sphere, we construct a KD-tree from them to efficiently search for the three nearest visual neighbors for each 4D Radar point on the sphere's surface.

Subsequently, we use a geometric constraint to estimate the scale factor. For each 4D Radar point \(p_i\), its three nearest visual neighbors in the camera coordinate system \({p_{a}, p_{b}, p_{c}}\), define a local physical plane. The normal vector for this plane is given by \(n=(p_a-p_b)\times(p_b-p_c)\). By projecting both the 4D Radar point and a visual neighbor onto this normal, we can compute the scale factor \(s_i\) for an individual 4D Radar point \(p_i\) as: \(s_i=\frac{n\cdot p_i}{n\cdot p_a}\). To ensure the validity of this estimation, we introduce two constraints. First, the method presupposes that the visual points lie on a common physical plane. This assumption is considered valid only when these points are sufficiently close to each other in their spherical projection and also have similar depth values in the reference depth map. Only then is the corresponding scale estimate considered reliable. Second, since depth estimates for dynamic objects are often inaccurate, dynamic 4D Radar points are excluded from this scale calculation. Finally, we statistically aggregate the scale factors computed from all valid 4D Radar points. The globally optimal scale is then determined using a robust histogram-based voting method.

\subsection{Joint Training Framework for 3DGS and VGPT}
\label{subsec:joint training}

To address the issue of inaccurate dynamic association, we introduce a direct and robust physical constraint for the deformation field by leveraging the radial velocity information from 4D Radar. Furthermore, to address the computational redundancy in existing methods \cite{chen2023periodic}, we depart from the strategy of training a global deformation field across the entire scene and instead propose a sparse, object-centric strategy, exclusively constructing and optimizing the deformation field within the dynamic regions identified by our segmentation module. Consequently, our model can be trained directly on high-resolution images, preserving the fine-grained details that are essential for high-fidelity scene reconstruction.
\begin{figure}[t]
    \centering
    \includegraphics[width=0.5\textwidth]{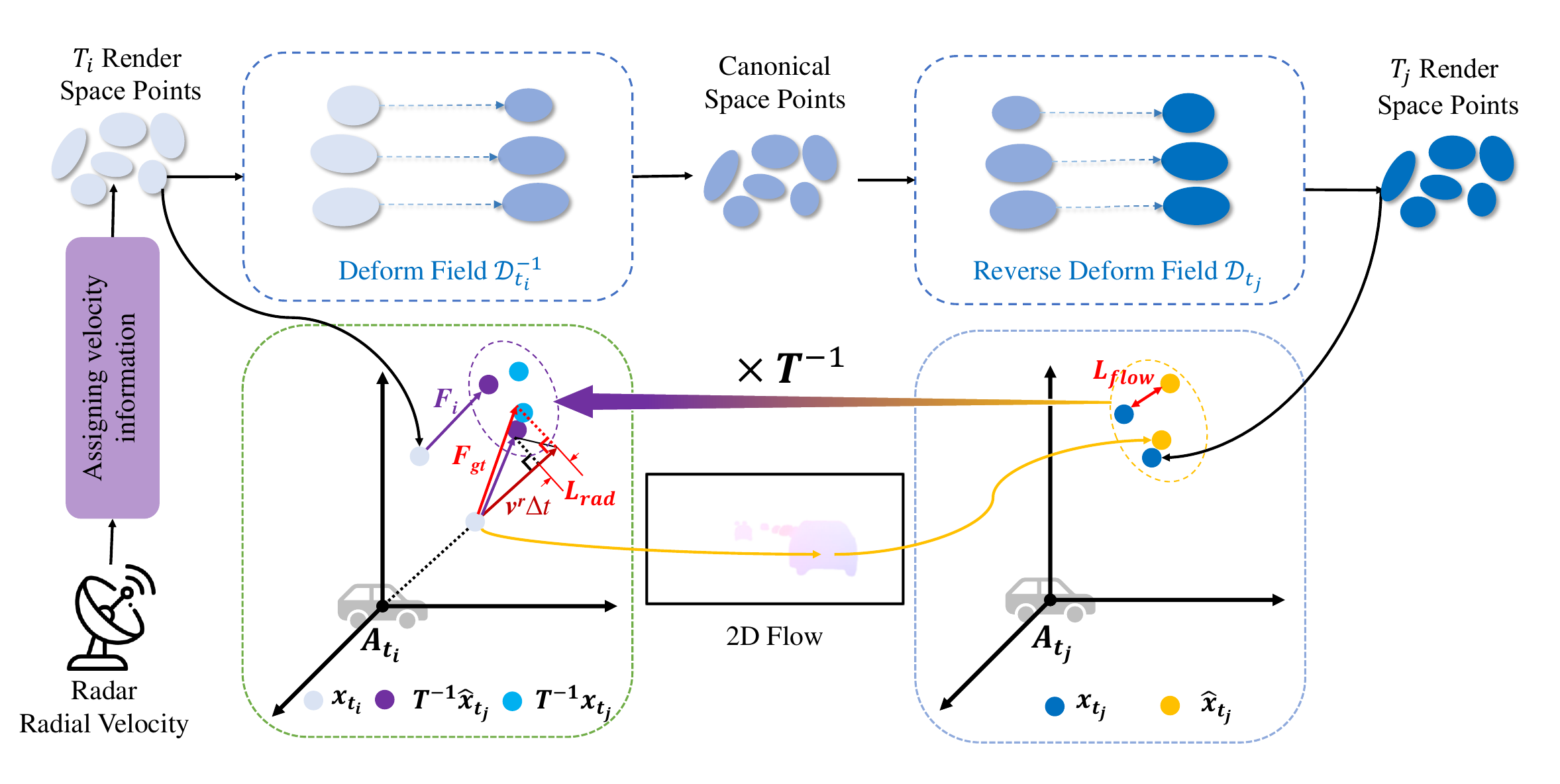}
    \vspace{-0.5cm}
    \caption{Overview of the VGPT Pipeline and its Dual Supervision. The pipeline (top) models motion by warping points from source time \(T_i\) to target time \(T_j\) via canonical space. This process is supervised by two signals (bottom): a geometric consistency loss derived from optical flow, and a direct physical constraint based on 4D Radar's radial velocity.}
    \vspace{-0.3cm}
    \label{pointtrack}
\end{figure}

\textbf{Deformation Field Modeling}.
For the spatio-temporal modeling of dynamic elements, we introduce a time-dependent deformation field \(\mathcal{D}_t:\mathbb R^3\to\mathbb R^3\), which warps point clouds from all timestamps into a common space to initialize a set of canonical Gaussian primitives. To render image at arbitrary time \(t\), these canonical Gaussians are transformed by the forward deformation field to their respective spatial locations for that time. They are then rendered from any given viewpoint using the standard 3DGS pipeline. 

Our deformation field \(\mathcal{D}_t\), inspired by Tracking everything \cite{wang2023tracking}, is implemented as an invertible Multilayer Perceptron network \cite{dinh2016density}. The network takes a 3D coordinate point \(x\in\mathbb R^3\) and a normalized timestamp \(t\in[0,1]\) as input to predict the point's new position \(x_t\in\mathbb R^3\). Owing to its invertible architecture, both the forward deformation \(\mathcal{D}_t\) and its inverse \(\mathcal{D}_t^{-1}\) can be analytically derived from a single forward pass, obviating the need for extra networks or iterative optimization. A single set of MLP weights is shared across all timestamps to ensure temporal continuity.
 
\textbf{Deformation Field Supervision}.
To optimize the deformation field, we use two complementary supervisory signals, as shown in Figure \ref{pointtrack}: (1) a pseudo-ground-truth 3D scene flow derived from optical flow,  (2) a direct physical constraint based on the radial velocities provided by the 4D Radar. 

\paragraph{Optical Flow-based 3D Scene Flow Supervision}
Given a pair of images at timestamps \(t_i\) and \(t_j\), we first employ a pre-trained optical flow model \cite{teed2020raft} to estimate the dense 2D optical flow field between them. As depicted in Figure \ref{pointtrack}, this 2D optical flow field establishes pixel-level correspondences for the dynamic Gaussian points across the two frames. Combined with the depth information provided by our monocular depth estimation network, these 2D correspondences are then lifted to sparse 3D correspondences between the two point clouds. This process generates a pseudo-ground-truth 3D scene flow. We leverage this 3D flow to supervise the learning of the deformation field \(\mathcal{D}\). Specifically, for any given dynamic point \(x_{t_i}\), its corresponding target position \(\hat{x}_{t_j}=\mathcal{D}_{t_j}\circ \mathcal{D}_{t_i}^{-1}(x_{t_i})\) at time \(t_j\) should be close to \(x_{t_j}\). The deformation field is then optimized by minimizing the following geometric consistency loss: 

\begin{equation}
    \mathcal{L}_{\text {flow }}=\sum\left\|\mathcal{D}_{t_j} \circ \mathcal{D}_{t_i}^{-1}\left(x_{t_i}\right)-x_{t_j}\right\|^2.
\end{equation}

\paragraph{Physical Constraints Based on 4D Radar Radial Velocity}

The ability of 4D Radar to directly measure the Radial Relative Velocity (RRV) of objects provides a physical prior that we leverage for direct supervision of our deformation field. First, we associate our segmented 3D dynamic Gaussians with the dynamic 4D Radar points at each timestamp using a K Nearest Neighbors (KNN) algorithm \cite{peterson2009k}. By setting K=1, we find the single nearest 4D Radar point for each Gaussian, from which the Gaussian inherits the ground-truth radial velocity measurement \(v^r\). Assuming uniform motion over a small time interval \(\Delta t\), the radial displacement between consecutive frames can be described by the geometric relationship shown in Figure \ref{pointtrack}: 

\begin{equation}
    v^r \Delta t=F_{g t}^{\top} \frac{(x_{t_i}-A_{t_i})}{\left\|x_{t_i}-A_{t_i}\right\|_2}.
\end{equation}

Here, \(A_{t_i}\) is the 4D Radar center, and \(F_{g t}=T^{-1}x_{t_j}-x_{t_{i}}\) is the flow vector of the ground truth scene for the point \(x_{t_i}\), where \(T^{-1}\) is the transformation matrix from the coordinate frame at time \(t_j\) to the frame at time \(t_i\). The projection of this vector onto the viewing direction must equal the radial displacement. Based on this physical constraint, we formulate a radial displacement loss \(\mathcal{L}_{rad}\), to directly supervise the radial component of the flow vector \(F_i\), generated by the deformation field. The predicted flow is defined as \(F_i = T^{-1}\cdot \hat{x}_{t_j} - x_{t_i}\). And the loss is: 

\begin{equation}
    \mathcal{L}_{rad}=\sum\left\|F_i^{\top} \frac{(x_{t_i}-A_{t_i})}{\left\|(x_{t_i}-A_{t_i})\right\|_2}-v^r \Delta t\right\|.
\end{equation}

 Despite certain unavoidable measurement errors, we empirically find that the RRV provides a strong supervisory signal, as demonstrated in Section \ref{subsec:ablation study}.

\begin{figure*}[t]
\centering
\includegraphics[width=1.0\textwidth]{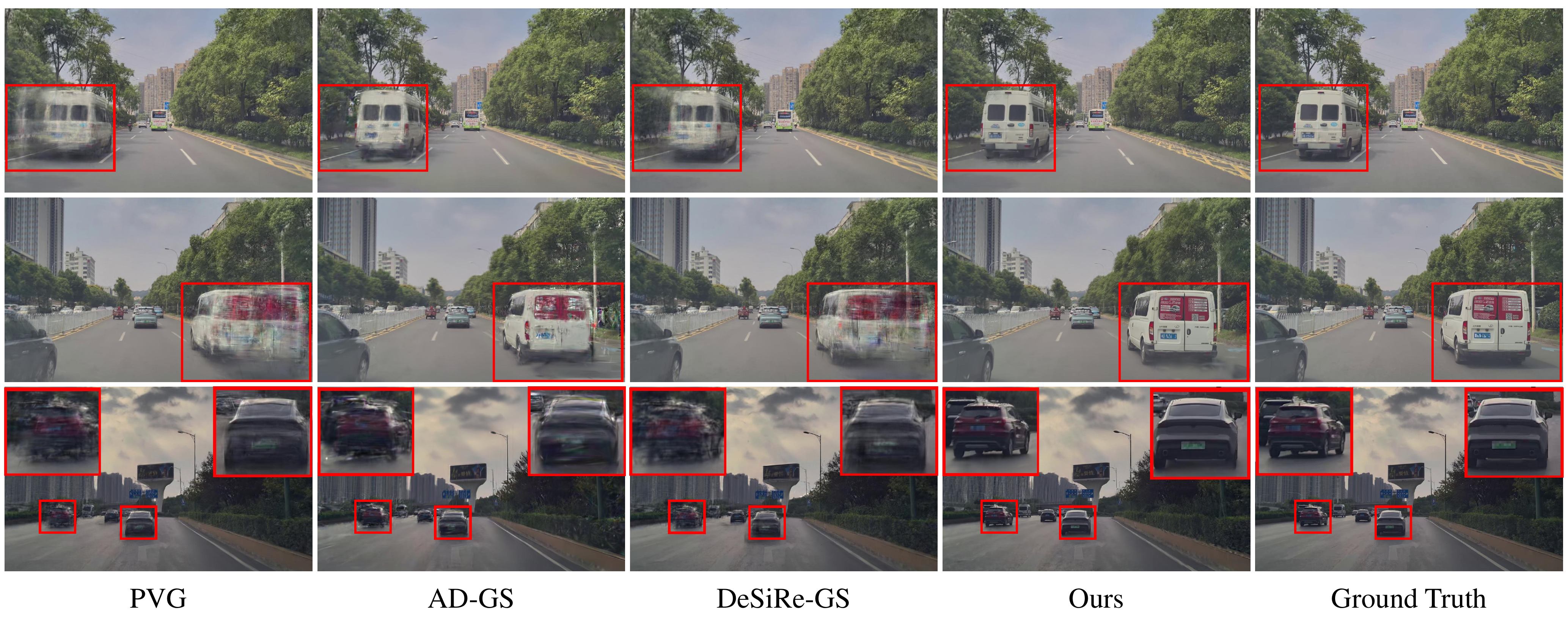}
\setlength{\abovecaptionskip}{-12pt}
\caption{Qualitative comparison of novel view synthesis. The red boxes highlight the reconstruction quality of dynamic vehicles. While prior methods suffer from severe artifacts like motion blur and ghosting, our method produces sharp and coherent reconstructions that are highly consistent with the ground truth.}
\label{comparison}
\end{figure*}

\begin{table*}[t]
    \centering
    \caption{\textbf{Quantitative comparison results on OmniHD-Scenes dataset}. The image resolution is $1920 \times 1080$. LPIPS uniformly adopts the VGG-Net. PSNR* for dynamic objects. \textbf{Bold}: Best. \underline{Underline}: Second Best.}
    \setlength{\tabcolsep}{4pt}
    \renewcommand{\arraystretch}{1.0} 
    \begin{tabularx}{\linewidth}{@{} lccccccccccccc @{}}
    \toprule
    \multirow{2}{*}{\textbf{Model}} & \multirow{2}{*}{\textbf{Venue}} & \multirow{2}{*}{\textbf{Type}} & \multirow{2}{*}{\textbf{BBox}} & \multirow{2}{*}{\textbf{3D Sensor}} & \multicolumn{4}{c}{\textbf{Scene Reconstruction}} & \multicolumn{4}{c}{\textbf{Novel View Synthesis}} \\
    \cmidrule(lr){6-9} \cmidrule(lr){10-13}
         &  & & & &  PSNR↑ & SSIM↑ & LPIPS↓ & PSNR*↑ & PSNR↑ & SSIM↑ & LPIPS↓ & PSNR*↑ \\ \midrule
        3D-GS \cite{kerbl20233d} & SIGGRAPH'23 & Static & $\times$ & - &  30.15 & 0.911 & 0.199 & 25.83  & 20.10 & 0.667 & 0.390 & 14.91 \\
        GaussianPro \cite{cheng2024gaussianpro} & ICML'24 & Static & $\times$ & - &  27.84  & 0.867 & 0.253  & 22.24 & 20.68 & 0.675 & 0.395 & 15.71 \\
        Octree-GS \cite{ren2024octree} & TPAMI'25 & Static & $\times$ & -  & 32.34 & 0.941 & 0.146 & 26.38 & 21.97 & 0.696 & 0.345 & 16.46 \\
        4D-GS \cite{wu20234d} & CVPR'24 & Dynamic & $\times$ & - & 29.97 & 0.874 & 0.242 &  26.75 & 20.63 & 0.691 & 0.396 & 15.83 \\
        Deformable-GS \cite{yang2024deformable} & CVPR'24 & Dynamic & $\times$ & - & 31.87 & 0.929 & 0.169 & 27.33 & 19.93 & 0.662 & 0.393 & 14.96 \\
        StreetGS \cite{yan2024street}  & ECCV'24 & Dynamic & $\checkmark$ & LiDAR  &33.58  &0.947  &0.139  &28.04  &26.55  &0.774 &0.270 &23.01 \\
        OmniRe \cite{chen2024omnire} & ICLR'25 & Dynamic & $\checkmark$ & LiDAR   &\underline{34.87}  &\underline{0.954}  &\underline{0.120}  &29.22  &\textbf{27.43}  &\underline{0.783} &\textbf{0.249} &\textbf{23.98}  \\
        S3Gaussian \cite{huang2024textit}   & arXiv'24 & Dynamic & $\times$ & LiDAR &34.77  &0.953  &0.121  &28.24  &20.24  &0.678 &0.357 &16.75  \\
        PVG \cite{chen2023periodic} & arXiv'23  & Dynamic & $\times$ & LiDAR &34.40  &0.952  &0.127  &28.28  &25.15  &0.767 &0.282 &21.48  \\
        DeSiRe-GS \cite{peng2025desire} & CVPR'25 & Dynamic & $\times$ & LiDAR  &33.98  &0.949  &0.135  &\underline{29.60}  &25.11  &0.767 &0.288 &21.46  \\
        AD-GS \cite{xu2025ad} & ICCV'25 & Dynamic & $\times$ & LiDAR  &32.51  &0.936  &0.153  &27.37  &25.42  &0.770 &0.277 &21.95  \\
        \midrule
        \textbf{Ours} & & Dynamic & $\times$ & 4D Radar & \textbf{34.96}   & \textbf{0.958} & \textbf{0.119} & \textbf{29.81} & \underline{26.68} & \textbf{0.790} & \underline{0.265} & \underline{23.33} \\ 
        \bottomrule
    \end{tabularx}
    \label{tab:com}
    \vspace{-0.2cm}
\end{table*}

\subsection{Training and Optimization Strategies}
\label{subsec:Training and Optimization Strategies}
\textbf{Gaussian Dropout Regularization}.
Potential inconsistencies in the 4D Radar-fused depth scales can lead to a critical reconstruction artifact: correctly positioned Gaussians are erroneously occluded by others. To mitigate this representational occlusion, we introduce a Gaussian dropout regularization method, inspired by the concept of Dropout in deep learning. During each training iteration, we stochastically set the opacity of a subset of Gaussian primitives to zero with a certain probability. This random dropping mechanism disrupts the established occlusion patterns in any given view, allowing Gaussians that would otherwise be occluded by foreground points to become exposed and receive supervision even under sparse view conditions. This approach effectively mitigates overfitting to specific training views and enhances the geometric consistency of novel view synthesis.

\textbf{Three-Stage Training Strategy}.
To ensure stable convergence and effective decoupling of our scene representation, we devise a three-stage training strategy. The first stage involves training the Gaussian network for the static background alongside the deformation field intended for the dynamic objects. In the second stage, we freeze the parameters of the static model and jointly train the dynamic Gaussians with the deformation field. The third stage is a global fine-tuning phase where all modules are unfrozen for end-to-end joint optimization of the static Gaussians, dynamic Gaussians, and the deformation field.

\textbf{Total Loss}.
The total training loss function is formulated as: 
\begin{equation}
    \mathcal{L}=(1-\lambda_r)\mathcal{L}_1+\lambda_r\mathcal{L}_{ssim}+\lambda_d\mathcal{L}_d+\lambda_{obj}\mathcal{L}_{obj}+\lambda_{sky}\mathcal{L}_{sky},
\end{equation}
where the \(\lambda\) terms are hyperparameters that weight each loss component. \(\mathcal{L}_1\) and \(\mathcal{L}_{ssim}\) are the L1 and SSIM losses on rendered images. \(\mathcal{L}_d\) is an inverse depth supervision loss following PVG \cite{chen2023periodic}, supervised by our scale-recovered depth maps. We adopt a mask constraint loss \(\mathcal{L}_{obj}=BCE(O_{dyn},M_{dyn})\) similar to that of StreetGS. This loss employs a binary cross-entropy term to align the rendered dynamic object mask \(O_{dyn}\), with the foreground mask \(M_{dyn}\), provided by our dynamic segmentation model. Following PVG, we employ a learnable environment map to represent the sky and apply a cross-entropy loss \(\mathcal{L}_{sky}=BCE(1-O_{sky},M_{sky})\) to enforce opacity in sky regions. Here, \(O_{sky}=\prod_{i=1}^N\left(1-\alpha_i\right)\) represents the accumulated transparency of all rendered Gaussians, and \(M_{sky}\) is sky mask predicted by Grounded-SAM \cite{ren2024grounded} model.

%% file: sec/04_exp.tex
\section{EXPERIMENTS}
\subsection{Experimental Setup}

\textbf{Datasets}.
As widely-used autonomous driving benchmarks such as the KITTI \cite{geiger2013vision}, Waymo \cite{sun2020scalability}, and nuScenes \cite{Caesar_Bankiti_Lang_Vora_Liong_Xu_Krishnan_Pan_Baldan_Beijbom_2020} datasets lack 4D Radar sensors, we evaluate our model on the OmniHD-Scenes \cite{zheng2024omnihd} dataset. This dataset is recorded at 10 Hz and is particularly suitable for our evaluation as it encompasses a rich variety of scenarios, ranging from rainy weather to clear days and from daytime to nighttime lighting, in addition to high-speed, low-speed, urban, and suburban settings. This variety is crucial for rigorously evaluating our model's robustness and its ability to generalize across different real-world conditions. From this dataset, we select 8 representative sequences for our experiments. Each sequence contains 50 frames, and we utilize data from the front-facing camera and front-facing 4D Radar. The camera resolution is downsampled to 1920×1080. For our train-test split, we assign every fourth frame of each sequence to the test set and use the remaining frames for training. 
\begin{figure}[tbp]
    \centering
    \includegraphics[width=0.48\textwidth]{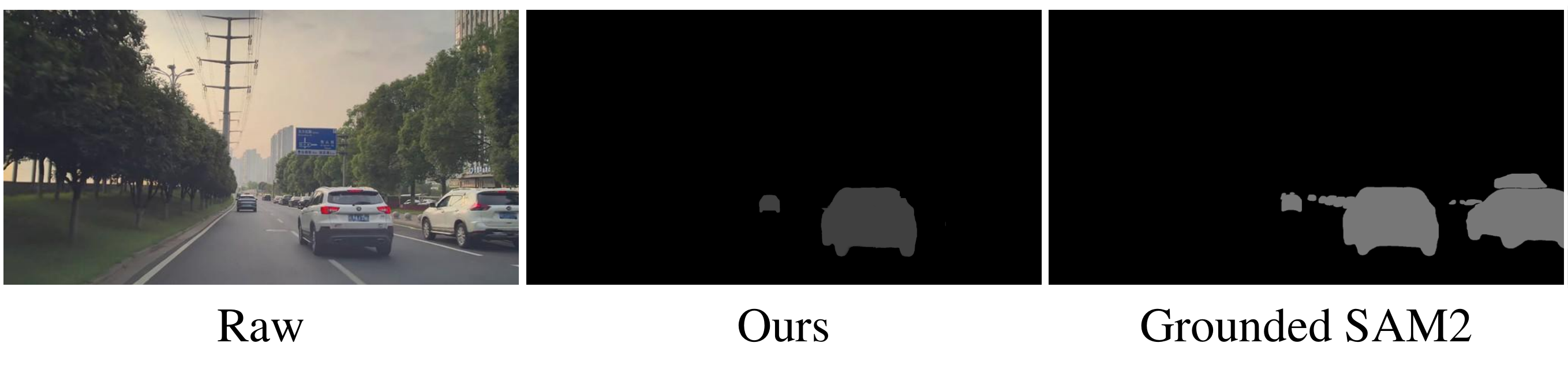}
    \setlength{\abovecaptionskip}{-12pt}
    \caption{Comparison of dynamic object segmentation under significant ego-motion.}
    \vspace{-0.1cm}
    \label{mask}
\end{figure}

\begin{table}[tbp]
\centering
\caption{Quantitative comparison of different mask types on \textbf{scene reconstruction} task.}
\label{tab:mask_comparison}
\setlength{\tabcolsep}{8pt} 
\renewcommand{\arraystretch}{1.2} 
\begin{tabular}{lcccc}
\toprule
Mask type & PSNR ↑ & SSIM ↑ & LPIPS ↓ & PSNR* ↑ \\
\midrule
Our & \textbf{34.96} & \textbf{0.958} & \textbf{0.119} & \textbf{29.81} \\
SAM2 \cite{ravi2024sam} & 34.52 & 0.952 & 0.120 & 28.73 \\
\bottomrule
\end{tabular}
\vspace{-0.1cm}
\end{table}
\textbf{Baselines}.
We evaluate our method against several state-of-the-art self-supervised approaches, including S3Gaussian \cite{huang2024textit}, PVG \cite{chen2023periodic}, DeSiRe-GS \cite{peng2025desire}, and AD-GS \cite{xu2025ad}. For a more comprehensive evaluation, we also compare our results with OmniRe \cite{chen2024omnire} and StreetGS \cite{yan2024street}, which are methods that require additional bounding box information for supervision. 

\textbf{Implementation Details}.
All experiments in this study were conducted on a single NVIDIA RTX 4090 GPU. 

Segmentation Model Training: We use dynamic 4D Radar points projected onto the image as prompts for Grounded-SAM \cite{ren2024grounded} to generate initial segmentation results for dynamic objects. From these, we manually select 11,973 high-quality results to serve as the ground truth for training. The input image resolution is 1920×1080, with a patch size of 256×256. The model is trained for 50 epochs using the Adam optimizer. The key hyperparameters are set as follows: a learning rate of 2e-4, a batch size of 6,  \(\beta_1=0.9\) and  \(\beta_2=0.999\). Total training time is approximately 12 hours. 

Gaussian Model Training: Our Gaussian model is trained for 30,000 iterations per sequence. The second stage of this training commences at 15,000 iterations, and the third stage begins at 20,000 iterations. The entire training process for each sequence takes approximately one hour to complete. We use learning rates similar to the original 3DGS implementation and set the loss weights as follows: \(\lambda_r = 0.2,\lambda_d = 0.1,\lambda_{sky} = 0.05,\lambda_{obj} = 0.05\). The \(\lambda_{obj}\) is only enabled in the third training stage. Sky masks are generated using Grounded-SAM \cite{ren2024grounded} with a 'sky' prompt. Regarding initialization, unlike the baseline methods which adhere to their respective protocols for supplementing background points, our approach directly initializes all Gaussians from the dense depth map. This eliminates the need for any additional point supplementation; the only filtering step is to discard points corresponding to the sky mask.

\subsection{Comparison}
\label{subsec:comparison}

\textbf{Comparison with Reconstructed Models}.
Following the evaluation protocol of PVG \cite{chen2023periodic}, we assess our method on two tasks: image reconstruction and novel view synthesis. The results are summarized in Table \ref{tab:com}, with the LPIPS metric computed using a VGG \cite{Simonyan_Zisserman_2015} backbone. In the self-supervised category, our method achieves SOTA performance across all rendering metrics for both reconstruction and synthesis tasks. The qualitative comparisons in Figure \ref{comparison} further corroborate these findings: previous methods like PVG and DeSiRe-GS exhibit severe ghosting and artifacts on dynamic vehicles, while AD-GS's reconstructions suffer from motion blur and noticeable noise. In contrast, by leveraging our effective point-wise tracking mechanism, our method generates sharp and coherent dynamic scenes, significantly outperforming existing self-supervised models in visual quality.

Notably, when compared to supervised methods that rely on additional bounding box annotations, our approach achieves highly competitive results without depending on any manual 3D labels. Its performance is on par with StreetGS \cite{yan2024street} and approaches that of OmniRe \cite{chen2024omnire}.

\begin{figure}[t]
    \centering
    \includegraphics[width=0.48\textwidth]{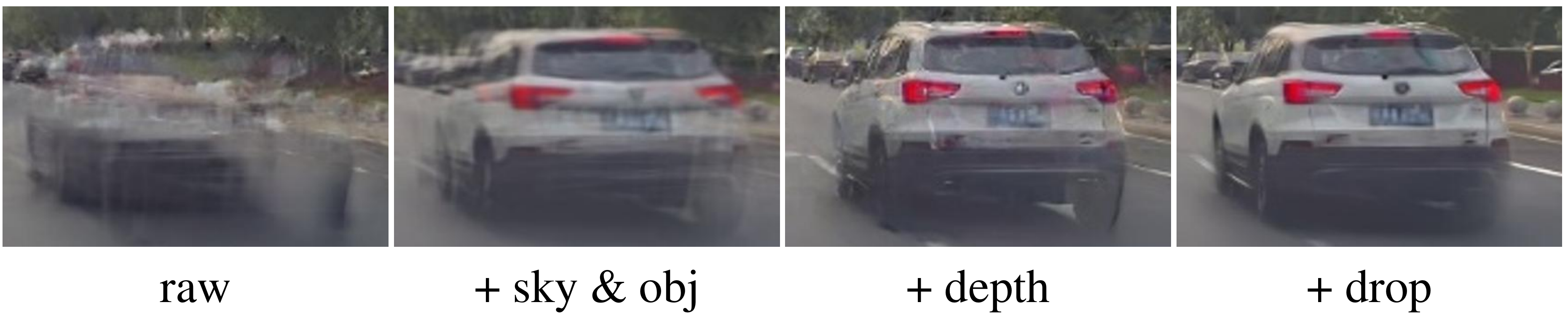}
    \setlength{\abovecaptionskip}{-12pt}
    \caption{Loss ablation by gradually adding the losses.}
    \label{loss}
\end{figure}

\begin{table}[t]
\centering
\caption{Ablation study on the contribution of each component on \textbf{novel view synthesis} task.}
\label{tab:ablation_final}
\setlength{\tabcolsep}{5pt}
\renewcommand{\arraystretch}{1.3}

\begin{tabular}{ccccccc}
\toprule
obj\&sky & depth & drop & PSNR ↑ & SSIM ↑ & LPIPS ↓ & PSNR* ↑ \\
\midrule
 & & & 23.46 & 0.750 & 0.299 & 17.44 \\
$\checkmark$ & & & 25.04 & 0.767 & 0.280 & 20.63 \\
$\checkmark$ & $\checkmark$ & & \underline{26.13} & \underline{0.783} & \underline{0.269} & \underline{21.70} \\
$\checkmark$ & $\checkmark$ & $\checkmark$ & \textbf{26.68} & \textbf{0.790} & \textbf{0.265} & \textbf{23.33} \\
\bottomrule
\end{tabular}
\vspace{-0.1cm}
\end{table}

\textbf{Comparison with Segmentation Models}.
Our segmentation method demonstrates superior robustness to significant ego-motion compared to vision-only approaches. As illustrated in Figure \ref{mask}, large camera displacements cause Grounded-SAM2 \cite{ravi2024sam} to misclassify static vehicles, leading to over-segmentation. In contrast, our method leverages the physical velocity priors from 4D Radar, demonstrating exceptional robustness to this issue. It successfully isolates only genuinely moving vehicles, thereby enabling more accurate parsing of the dynamic scene.This improved accuracy is quantitatively validated by the results presented in Table \ref{tab:mask_comparison}.

\subsection{Ablation Study}
\label{subsec:ablation study}
\textbf{Ablation of Gaussian Loss}.
We conduct an ablation study to validate the effectiveness of our model's components and their contributions to the novel view synthesis task. This study involves the progressive incorporation of several components atop a baseline model: the combined object and sky mask losses \(\mathcal{L}_{\text{obj}} + \mathcal{L}_{\text{sky}}\), the depth supervision loss \(\mathcal{L}_d\), and our dropout regularization. As demonstrated by the results in Table \ref{tab:ablation_final} and Figure \ref{loss}, each component provides a distinct benefit: the 2D supervision from the mask losses is crucial for facilitating a clean spatial separation of object and background Gaussians; the 3D geometric constraints from the depth loss further enhance the accuracy of structural reconstruction; and finally, the Gaussian dropout regularization improves geometric robustness by effectively suppressing artifacts and ensuring greater cross-view consistency.
\begin{figure}[t]
\centering
\includegraphics[width=0.48\textwidth]{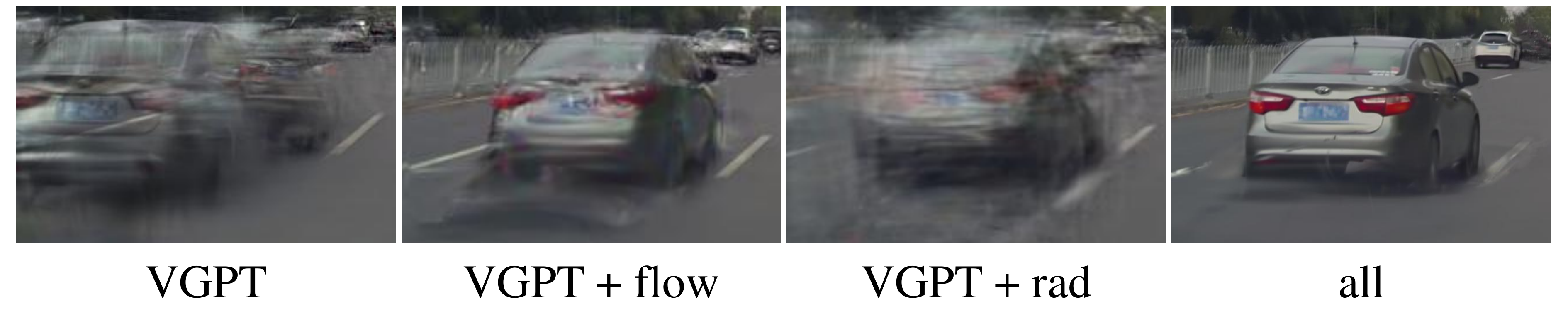}
\setlength{\abovecaptionskip}{-12pt}
\caption{Ablation Study on the supervision for the VGPT Model.}
\label{track}
\end{figure}
\begin{table}[t]
\centering
\caption{Ablation study on the supervision of our VGPT model on \textbf{novel view synthesis} task.}
\label{tab:pointtrack_ablation}
\setlength{\tabcolsep}{6pt} 
\renewcommand{\arraystretch}{1.2} 
\begin{tabular}{lcccc}
\toprule
 & PSNR ↑ & SSIM ↑ & LPIPS ↓ & PSNR* ↑ \\
\midrule
VGPT & 25.02 & 0.762 & 0.305 & 18.38 \\
VGPT+flow & \underline{25.97} & \underline{0.775} & \underline{0.273} & \underline{22.29} \\
VGPT+rad & 25.58 & 0.770 & 0.285 & 20.93 \\
all & \textbf{26.68} & \textbf{0.790} & \textbf{0.265} & \textbf{23.33} \\
\bottomrule
\end{tabular}
\vspace{-0.1cm}
\end{table}

\textbf{Ablation of VGPT Loss}.
To validate the effectiveness of the deformation field supervision in our proposed VGPT model, we incrementally add the network, \(\mathcal{L}_{flow}\) and \(\mathcal{L}_{rad}\), corresponding to 'track', 'flow', and 'rad', respectively. As demonstrated by the results in Table \ref{tab:pointtrack_ablation} and Figure \ref{track}, without the VGPT model, the model fails to account for the object's motion, resulting in extreme motion blur that renders the vehicle's form nearly indecipherable. Introducing the tracking network supervised solely by the optical flow-based loss substantially reduces this blur and recovers the vehicle's general shape, yet significant ghosting artifacts and a lack of sharp details remain. Conversely, using only the 4D Radar-based radial displacement loss is insufficient to constrain the complex 3D motion, leading to severe geometric distortion and an uninterpretable reconstruction. By combining both supervisory signals, our full model synergistically leverages the dense 2D guidance from optical flow and the precise physical constraints from 4D Radar to generate a sharp, coherent, and artifact-free reconstruction.

\begin{figure}[t]
    \centering
    \includegraphics[width=0.485\textwidth]{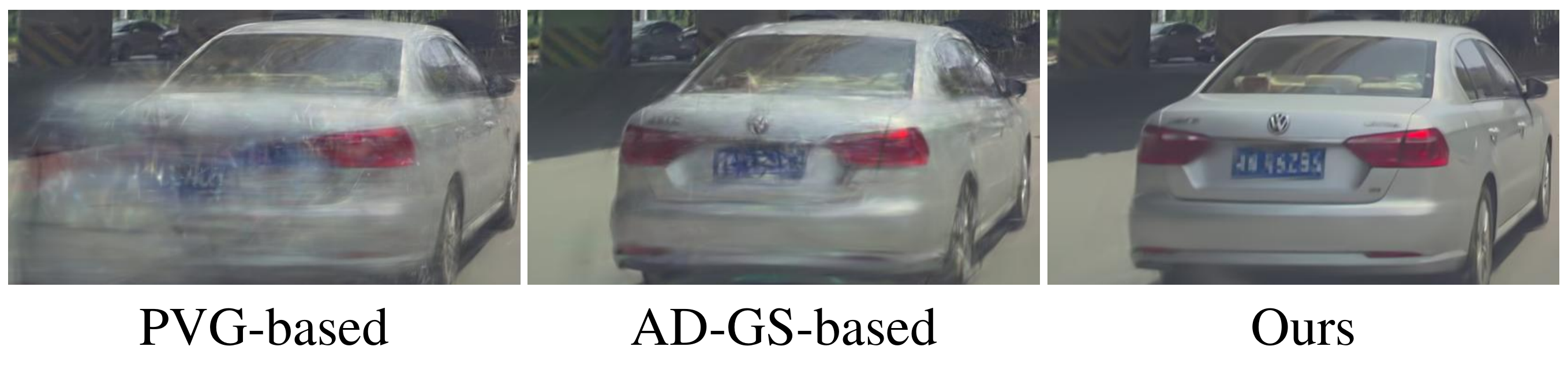}
    \setlength{\abovecaptionskip}{-12pt}
    \caption{Qualitative comparison of different association strategies.}
    \label{model}
\end{figure}

\begin{table}[t]
\centering
\caption{Quantitative comparison with baseline methods on \textbf{novel view synthesis} task (evaluated on 4 sequences).}
\label{tab:baseline_comparison}
\setlength{\tabcolsep}{8pt} 
\renewcommand{\arraystretch}{1.2} 

\begin{tabular}{lcccc}
\toprule
 & PSNR ↑ & SSIM ↑ & LPIPS ↓ & PSNR* ↑ \\
\midrule
PVG-based & 25.92 & 0.800 & 0.240 & 19.03 \\
AD-GS-based & \underline{27.04} & \underline{0.808} & \underline{0.231} & \underline{21.48} \\
Our & \textbf{27.68} & \textbf{0.820} & \textbf{0.225} & \textbf{23.74} \\
\bottomrule
\end{tabular}
\end{table}

\textbf{Ablation of Dynamic Associations}.
We compare the association strategies of PVG \cite{chen2023periodic} and AD-GS \cite{xu2025ad}, conducting all experiments across four representative sequences under identical initialization and masking conditions to ensure a fair comparison. The results are presented in Table \ref{tab:baseline_comparison} and Figure \ref{model}. Our findings reveal that in reconstructing dynamic driving scenes, the novel view synthesis results from the two methods exhibit distinct characteristics and trade-offs. PVG's reliance on a periodic vibration model makes it susceptible to correspondence errors, which in turn introduces dynamic artifacts. In contrast, AD-GS employs B-splines to enforce the temporal smoothness of keypoint trajectories. While this approach effectively mitigates the issue of association failure, its strong smoothing prior hinders the precise capture of subtle dynamics, leading to a loss of fine detail in the final reconstruction.

\begin{table}[h]
\centering
\caption{Quantitative comparison of initialization methods on \textbf{scene reconstruction} task, evaluated within the PVG framework.}
\label{tab:init_comparison}
\setlength{\tabcolsep}{12pt} 
\renewcommand{\arraystretch}{1.2} 

\begin{tabular}{lccc}
\toprule
\textbf{initialize type} & PSNR ↑ & SSIM ↑ & LPIPS ↓ \\
\midrule
4D Radar & \textbf{34.4011} & \underline{0.9517} & \underline{0.1274} \\
LiDAR & \underline{34.3979} & \textbf{0.9521} & \textbf{0.1268} \\
\bottomrule
\end{tabular}
\vspace{-0.1cm}
\end{table}
\textbf{Ablation of Sensor}.
To fairly evaluate and compare the performance differences between 4D Radar and LiDAR as data sources for initialization, we conducted a set of controlled experiments within the PVG \cite{chen2023periodic} framework. While keeping other experimental variables, such as the segmentation masks, strictly consistent, we initialized the Gaussian primitives using two different approaches: one based on pure LiDAR point clouds, and the other using our proposed method which fuses 4D Radar data with monocular depth estimates. A quantitative comparison of the two initialization strategies is presented in Table \ref{tab:init_comparison}. As the results indicate, 4D Radar-based initialization achieves a level of accuracy comparable to that of the denser LiDAR point clouds.

%% file: sec/05_conclu.tex
\section{CONCLUSIONS}
We present a self-supervised 3D Gaussian Splatting framework that, for the first time, leverages 4D Radar priors for high-fidelity dynamic driving scene reconstruction without manual annotations. Our method uses 4D Radar velocity for robust dynamic segmentation and its radial velocity as a direct physical constraint to supervise a deformation field alongside optical flow for accurate motion modeling. Experiments demonstrate state-of-the-art performance among self-supervised dynamic driving scene reconstruction methods.